# Synthetic Data in Marketing Research: How to Evaluate and When to Trust

Oded Netzer[+] and Rajan Sambandam[++]

Debate over synthetic data in marketing research has polarized into claims that LLMs make human respondents obsolete and calls to shun them entirely. We argue that both positions obscure the more useful question: not *whether* synthetic respondents work, but *when*. Building on Brand, Israeli, and Ngwe (2026), we make three contributions. First, we distinguish three types of synthetic data: ungrounded LLM responses, segment-level personas, and individual-level digital twins, and map each to the decisions it can support. Second, we develop a taxonomy of four families of accuracy measures and suggest that the wide discrepancies in reported twin accuracy (from near-perfect to near-chance) could largely reflect differences in what is being measured rather than in method quality. Aggregate measures often perform reasonably well even when little information is provided to the LLM and can mask the complete absence of respondent-level differentiation. Third, we introduce the *forgotten question* problem (a question omitted from a fielded study) as a promising setting for twin-based augmentation to existing studies. We propose an ex-ante answerability diagnostic requiring no ground-truth data: the $R^2$ of a random forest predicting twin outputs from twin training data. Across 108 attitude questions from a nationally representative survey (N = 3,063), an $R^2 > 0.7$ screen raises mean twin–human individual-level correlation by 15% and reduces the share of poorly answered questions from 25.9% to only 4.3%. Embedding similarity provides a correlated but weaker screen, and experienced-researcher judgment was comparably informative but validated on fewer questions.

**Acknowledgement:** The authors would like to thank TRC Insights for help with data collection, analysis, and valuable discussions, especially Cory Boyko, Marc Hershey, Westley Ritz, and Kevin Dona. We also wish to thank Malek Ben Sliman, George Gui, Leonard Kinzinger, Moses Miller, Yuchen Qiu, and Olivier Toubia for the helpful comments and suggestions.
+ onetzer@gsb.columbia.edu. Arthur J. Samberg Professor of Business, Columbia Business School
++ rsambandam@trcinsights.com. President, TRC Insights

## 1. Introduction

The year is 2016. A health insurance executive needs to understand how potential customers will choose health plans on the exchanges that opened after the passage of the Affordable Care Act. She approaches an experienced market researcher for help in conducting a study to gather the relevant input for the business decision. The researcher understands that a conjoint analysis is needed and has done enough of them in the healthcare category to know the general contour of the outcome. If pressed, the researcher could probably predict most of the findings (barring novel attributes or surprising results) and even start drafting the final report on the very first day of the project. But he won't, for at least a couple of good reasons. One, it hardly makes business sense to simply offer the known result and conduct a study simply for the unknown sliver (even if such a clean bifurcation were possible). Two, the business executive would be unwilling to stand behind such a result without data to back it up, regardless of how credentialed the market researcher is.

Now let's fast forward to 2026. The same problem can be approached in a much different light. LLMs trained on much of the knowledge base of the Internet are easily available and are being used at scale. Going back to our expert marketing researcher, one could argue that given the wealth of marketing research projects and conjoint analysis studies in its training data, LLMs' knowledge equals or exceeds that of experts, especially in its comprehensiveness. Should that change the decision of whether we should use the expert (in this case, an LLM) input to answer the client's research questions? While it is still not clear that we want to use an LLM as the sole source of an answer to the client, throwing away this wealth of knowledge, at least in some cases, seems wrong as well.

What the client is paying for is not the 80% of the study results that are predictable a priori by the seasoned researcher or AI, such as the price coefficient being negative or the price elasticity being larger than advertising elasticity. In fact, it is these predictable results that will help the client gain confidence in the study's findings. The reality is that if AI were to give us surprising results, skepticism would be warranted given the danger of hallucinations. The client is paying for the 10-20% of "real" insights or surprises that were not known before the study started.[1]

Our argument here is that AI, specifically synthetic data, can be a very useful aid for market researchers. But in the current climate of polarization over the use of synthetic data, a nuanced perspective is needed to delineate how and when synthetic data should be used.

---

[1] https://www.nytimes.com/2024/12/23/science/ai-hallucinations-science.html

As the adoption of generative AI increases, we have seen polarization in the academic literature and public media on the use of generative AI in marketing research. For example, the synthetic-sample company Aaru received a very favorable review from the *Wall Street Journal*[2], and, at the same time, drew industry pushback.[3] At one pole, Argyle et al. (2023) introduced the notion of "algorithmic fidelity" and showed that even early models such as GPT-3, conditioned on sociodemographic backstories, could generate "silicon samples" whose response distributions closely track those of human subgroups. Horton (2023) further proposed treating LLMs as homo silicus endowed economic agents who can be run through classic experiments essentially for free. Dillion et al. (2023) reported a correlation of r = .95 between GPT-3.5 and average human moral judgments, and Park et al. (2026) report that agents grounded in two-hour interviews with over 1,000 Americans recover 83% of those individuals' own test–retest consistency on held-out survey items. Read quickly, this body of work licenses a strong conclusion: that human respondents are, at the margin, becoming optional.

At the other pole sits an equally forceful literature. Bisbee et al. (2024) document that silicon samples are under-dispersed, highly sensitive to prompt wording, and unstable across model versions. Wang et al. (2025) argue that LLM surrogates do not merely approximate demographic groups but flatten and misportray them. Harding et al. (2024) and Lin (2025) make the categorical case against substituting human respondents for synthetic data on methodological and conceptual grounds. Written by industry partners, Morris et al. (2025) conclude that synthetic data *"should not be used"* as a substitute for public opinion and survey data.

Both poles have incentives. The optimists often have strong commercial incentives. Vendors currently advertise the replacement of anywhere from 15% to 85% of human respondents, and industry forecasts project that synthetic data will "completely overshadow" real data in AI development by 2030.[4] But it is worth noting that the pessimists have commercial interests too. A great deal of the loudest skepticism originates with firms whose business is human sample, and their position tends toward a conservatism that denies the real value this technology provides. Academic research or industry-academia collaborations, such as the present paper, which are likely to have weaker commercial incentives, can serve as the disinterested referee, and Brand, Israeli, and Ngwe

[2] https://www.wsj.com/business/ai-startup-aaru-young-founders-35da7f87
[3] https://www.thevoiceofuser.com/aaa-billion-dollar-ai-startup-is-selling-you-a-survey-the-wall-street-journal-wrote-a-love-letter-about-it/
[4] https://www.linkedin.com/pulse/synthetic-data-generation-overcoming-scarcity-ai-model-training-q0soc/

(2026) provide a model of what that role looks like. Transparent benchmarks, a mixed verdict, and no incentive to round the answer toward either pole.

We think that much nuance is lost in this argument and propose that synthetic data, if used selectively and appropriately, can provide value. As we have experienced in many areas related to generative AI, the real issue is that the barrier to entry is low while the barrier to excellence is high (Sambandam and Netzer 2025). This is particularly true for the use of synthetic data. Hence, it is easy to hawk but hard to validate (but not impossible, as we will demonstrate).

We can start by outlining what it is, exactly, that we mean by synthetic data.

## Types of synthetic data

One could categorize the different forms of synthetic data into three types at varying levels of aggregation:

- *Ungrounded LLM responses* (Type 1) – Simply using LLMs with no additional human grounding is at the highest level of aggregation.
- *Segment-level personas* (Type 2) – Using digital personas to represent segments for chatbot interaction, generally based on segment-level human data.
- *Individual-level digital twins* (Type 3) – Digital twins, which are based on and validated on individual humans, provide the most granular form of synthetic data.

Each type has its uses. For business decisions that are currently made in real time based on gut, with little to no data, aggregate LLM responses with no input data (Type 1) may be helpful. It can complement, augment, or even replace such decision-making. This approach can also be used in the earlier stages of the marketing research process (Arora et al. 2025) or for qualitative marketing research (Korst et al. 2026). Just having access to a good LLM is a good starting point. However, this type of data can be useful even for more complex and complete forms of market research. In fact, the Brand, Israeli, and Ngwe (2026) article uses this type of data and clearly provides reasonable results (perhaps even better than what the experienced market researcher could consistently provide).

Segment-level personas (Type 2) data provide a finer cut, allowing for more nuanced testing. However, it requires more time and expertise to develop. A good segmentation study with an appropriate human sample would first need to be conducted, followed by the development of personas that can reasonably represent the centroids of those segments. A difficulty with such personas is that there is no individual-level data to contrast the personas with, making it difficult to validate. However, when done well, studies like claims testing across different segments could be simulated. Personas are particularly useful for qualitative work and communicating the research result.

Individual-level digital twins (Type 3) data are the highest standard. Significant amounts of relevant human data are collected to build a twin of each respondent. They can be combined to provide segment-level views or aggregated population-level views. As we discussed earlier, even with a large amount of testing required beyond the other types, substantial variation remains in twin performance depending on how well they were built and what they are asked. But if done well, this can provide value. As with other contexts, such as conjoint analysis, individual-level data are useful for any level of aggregation as well as to capture heterogeneity.

For the rest of this paper, we primarily focus on individual-level digital twins as this is the most rigorous form of synthetic data, and present some empirical results of our twin testing to demonstrate the possible value of digital twins.

## 2. The Shades of Gray in the Use of LLMs for Market Research: Learnings from Brand, Israeli, and Ngwe (2026)

Brand, Israeli, and Ngwe (2026; hereafter BIN) make a seminal contribution to what is becoming one of the most consequential questions in marketing research: whether and how large language models can stand in for, or augment, human respondents. It is good to see this work in print, as it serves as a crucial first step towards tackling this important problem. BIN have taken on a problem that practitioners are already confronting perhaps with more enthusiasm than evidence, and they bring to it the discipline of carefully designed conjoint studies, real human benchmarks, and a transparent accounting of where the approach succeeds and where it does not. Establishing that kind of rigorous baseline is what a fast-moving field with varying levels of incentives needs.

One of the most important aspects of the BIN paper is its willingness to deliver a mixed verdict. The authors show that GPT's out-of-the-box estimates of willingness-to-pay are sometimes in the right ballpark but are frequently unreliable, overestimating WTP for fluoride threefold, assigning the wrong sign to brand preferences and to new toothpaste flavors, and varying substantially across model versions. Crucially, they emphasize that a researcher lacking a human benchmark would have had no way to know in advance which estimates to trust.

The reality is that AI's ability to mimic human preferences is limited and, to its credit, this paper clearly makes that point. Academic publishing tends to reward clean, definitive answers, so it's hard to write and publish a paper whose central message is "it depends, and often we cannot tell when." But that is the state of the world of synthetic data in marketing research in 2026: the answer lives in shades of gray, and we frequently do not even know whether a given shade is closer to white or black. The BIN paper should be lauded for resisting the temptation to oversell.

Following Horton (2023), BIN note that LLM training corpora may be closer to revealed preferences (actual purchases) while conjoint surveys elicit stated preferences. Under that lens, some apparent "failures", like GPT's strong income conditioning on expensive electronics, may not reflect the model being wrong, but rather the model measuring a related but distinct preference object. This reframes the whole evaluation exercise in a productive way: the right benchmark may not be obvious, and divergence is not automatically error. Future research could further explore this point.

That said, some of the limitations of the BIN paper (most of them highlighted by the authors) are worth noting as key lessons for the future. The paper mostly rests on GPT-3.5 Turbo. It conducts some robustness checks using GPT-4o and demonstrates a meaningful difference in the estimates produced by the newer model. This highlights the rapidly evolving nature of the field and the need to re-evaluate the usefulness of AI for marketing research as this technology evolves. The BIN paper also highlights the limitations of LLMs in quantifying uncertainty and capturing cross-customer heterogeneity.

The authors correctly explain why conventional standard errors are misleading in the case of synthetic data (particularly Types 1 and 2) and why repeated queries do not mimic true respondent heterogeneity. The number of LLM queries is a design choice rather than a sampling constraint and conventional standard errors would shrink toward zero with more queries, creating a dangerous illusion of precision around estimates that may simply be wrong.

The most encouraging and practically actionable finding is the paper's demonstration that fine-tuning on prior human survey data can extend an LLM's usefulness to product features that were never in the fine-tuning data. For instance, fine-tuning with conventional toothpaste data flips the (incorrect) signs on willingness-to-pay for novel toothpaste flavors (cucumber and pancake) to match the human direction, and fine-tuning with laptop data brings the model's estimate for a built-in projector (badly overstated at baseline) into closer alignment with the human benchmark. One way to read this result is as a problem of distribution shift. Mapping an individual profile onto that individual’s responses is a task that sits largely outside what current models were trained to do, and fine-tuning on human survey data from the same category supplies the mapping that is missing, but only locally, for questions near the data supplied. Within-category extrapolation to a new attribute works because the degree of distributional shift is likely to be low. Across-category transfer, laptops to tablets, does not. And recovery of demographic heterogeneity, which asks the model to extrapolate along a dimension the fine-tuning data barely identifies, fails as well.

This within-category, new-attribute extrapolation is where the approach shows its clearest promise and points to a potential direction: a researcher holding survey data on one set of features may be able to generate credible preliminary estimates for adjacent new features before committing to a full new study. The question then is, when can a question or a feature be considered sufficiently “adjacent” for an LLM to provide a reliable answer? We will unpack this question and provide further empirical evidence in Section 4.

## 3. How Do We Measure Digital Twins’ "Accuracy"?

Part of the polarization around synthetic data and digital twins for marketing research stems from the fact that some academic and professional studies report near-perfect accuracy levels, while others report near-random accuracy levels. A reader of this literature could reasonably conclude that digital twins are 95% accurate, 75% accurate, or barely better than chance, and each conclusion has published support.

For example, Dillion et al. (2023) find a near-perfect correlation of r = .95 between GPT-3.5's ratings of 464 moral scenarios and average human judgments. Digital twins built from rich individual-level data were able to answer never seen before questions at roughly 72–76% raw accuracy measured (using hit rates for binary responses or distance measures for continuous responses; Toubia et al. 2025; Ye and Yoga Narasimhan 2026; Peng et al. 2026). On the other hand, similar twins based on people’s responses to 500 questions outperformed a naive prompt by barely more than one percentage point in accuracy (0.748 vs. 0.734, against a floor of 0.629 for uniformly random responding; Peng et al. 2026).

We argue that these discrepancies reflect differences in *what is being measured* far more than differences in method quality. Consider three prediction targets in ascending order of difficulty. The easiest is a population average, such as the mean rating of a set of moral scenarios (Dillion et al. 2023). Next is the correlation across questions for each respondent (Park et al. 2026), which is easier to predict than the correlation across respondents on a specific construct (Peng et al. 2026). Averaging across people removes the need to detect individual differences. These different studies report startlingly different accuracy levels, but they also use different measures of accuracy. The field lacks a shared standard for determining which accuracy measure is appropriate for a given claim. In this section we define a taxonomy of accuracy measures and map measures to the decisions academics and practitioners actually face.

### 3.1 A taxonomy of accuracy measures

Let *Y(q,i)* denote person *i*'s response to question q, and *Ŷ(q,i)* the response of the synthetic agent constructed to represent person *i*. We categorize the measures in the literature into four families, ordered here from least to most demanding of the data.[5] Three questions generate these families and the order in which we present them. First, what does the measure require: matched individual records, or only question-level summaries? Second, what do we aggregate on: across people within a question, or across questions? Third, does it score the level of a response or only its ordering? The families are therefore not a list of alternatives but a hierarchy of what a measure demands of the data.

*Family 1: Cross-question correspondence.*

Correlate question-level summaries, typically means, *across* questions, scenarios, or studies, or pool all (person, question) observations into a single correlation. No matching and no person-level data are required at all, only item-level summaries.

*Family 2: Question-level distributional comparisons.*

Compare the aggregate distribution of synthetic and human responses question by question. Differences in means (e.g., Glass's Δ), ratios of standard deviations, choice shares, full-distribution distances such as the Wasserstein (earth-mover's) distance, and derived aggregate estimands such as logit partworths, willingness-to-pay, and demand curves have been used here. No one-to-one matching between humans and twins is required. This is BIN's design: repeated LLM queries generate a synthetic sample whose implied WTP and choice shares are compared to human conjoint benchmarks by sign, magnitude, and ordering.

*Family 3: Matched individual-level accuracy.*

For each question and each person, compare their response to the response of their own twin:

$$Accuracy(q) = \frac{\sum_i \left[1 - \frac{|Y(q,i) - \hat{Y}(q,i)|}{Range(q)}\right]}{N} \text{ or } Accuracy(q) = \frac{\sum_i I[Y(q,i) = \hat{Y}(q,i)]}{N},$$

[5] Within each family, different measures of accuracy, such as mean absolute deviation, mean squared error or hit rates can be used.

where $q$ is a question, $i$ is a respondent out of a total of N respondents, $Y(q,i)$ is the response of respondent $i$ to question $q$, and $\hat{Y}(q,i)$ is the corresponding digital twin prediction. The first term is individual-level accuracy for continuous variables, and the second is a hit rate measure for discrete variables.[6] These measures can then be aggregated across questions.

This family measures a one-to-one mapping between real and synthetic respondents (Toubia et al. 2025; Peng et al. 2026; Ye and Yoga Narasimhan 2026). It asks: does the twin reproduce *this person's* answer to *this specific question*?

*Family 4a: Within-person, across-questions correlation.*

For each individual, the correlation between their responses to a set of questions and their digital twin responses can be calculated by:

$r(i) = correlation_q(Y(q,i), \hat{Y}(q,i))$.

This asks a different question for each person: how correlated are their true and twin answers across questions? This measure can then be aggregated across individuals. This measure is a combination of individual-level responses and aggregate responses because a good aggregate response (e.g., question-level population means) with no heterogeneity can still provide high correlations. Thus, this correlation measure is likely to be easier for the digital twin to predict relative to the within-question correlation discussed next.

*Family 4b: Within-question, across-person correlation.*

For each question, the correlation between the responses of the original respondents and their digital twins can be measured by:

$r(q) = correlation_i(Y(q,i), \hat{Y}(q,i))$.

Like Family 3, this measure also requires matching, but asks a different question: does the twin *sort* people correctly in terms of who scores higher than whom, even if the predictions are systematically shifted? This measure can then be aggregated across questions. Note that neither levels nor ordering detects a twin whose answers are correctly ranked but too tightly spread. Agreement measures such as Lin's (1989) concordance correlation penalize both level shifts and compressed spread, sitting conceptually between Families 3 and 4b.

Family 3 and Family 4b differ in terms of the use of information-free knowledge from the LLM training data. Consider two questions on a seven-point scale, one with a human mean of 5 and one with a mean of 2, each with a standard deviation of about 1 across respondents. An LLM that knows from its training corpus roughly where each question sits, will perform well on Family 3, while its Family 4b correlation within each question may still be low. Thus, part of Family 3 and Family 4a accuracy can come from what the model knows about the population, not from anything it knows about the respondent.

[6] These accuracy measures could be further adjusted for ordinal or multinomial responses.

The families differ sharply in cost, which explains why Family 1 dominates the early literature and Families 3 and 4b remain rare but are most likely to appear in top peer-reviewed academic publications (e.g., Toubia et al. 2025; Peng et al. 2026). However, importantly, these measures differ sharply in what they measure, what they can support, and the impression they may give of the level of accuracy. Kinzinger and Hartmann (2026) demonstrate this directly. Scoring 2.1 million twin responses across a grid of construction choices on held-out panel data, they find that enabling an explicit reasoning mode raises rank-order correlation (Family 4b) while leaving hit-rate accuracy unchanged (Family 3).

Individual-level accuracy (Families 3 and 4b) dominates the others because information flows one way. Get individual responses right, and every aggregate follows. On the other hand, matching aggregates implies nothing about individuals. This is exactly the rationale for estimating individual-level partworths in conjoint. Using individual-level partworths, you can target individuals, cluster into segments, and simulate aggregate share, whereas pooled partworths can never be disaggregated back.

Formally, if twins are compared to their humans' responses and mean $|\hat{Y}(q,i) - Y(q,i)| \leq \varepsilon$ with $\varepsilon$ well below the between-person standard deviation of $Y(q,i)$, every aggregate inherits that level of accuracy.

### 3.2 Aspects to consider across accuracy measures

First, and most importantly, correlations computed across questions rather than across people within a question conflate aggregate item effects with person effects. The variance of $Y(q,i)$ can be decomposed into a *between-question component* (items could elicit high or low means due to item popularity, difficulty, valence, scale polarity, etc.) and a *within-question, between-person component* (real heterogeneity in who answers what).

The between-question component is much easier for LLMs to predict based on their training corpus without knowing anything about any individual. A simulator that returns the *identical* answer for every synthetic respondent, with pure population priors and zero individuation, can approach a cross-item correlation of 1. Two further properties make Family 1 statistics hard to read. A correlation is invariant to affine rescaling, so it is silent on magnitude: Ashokkumar et al. (2026) predict treatment effects across 70 experiments with a correlation of 0.85, but those predictions are roughly twice the true effect sizes. Additionally, when looking at the different treatment arms within an experiment, these correlations drop to 0.39. These may be legitimate answers to item-level and effect-level questions. Inflation occurs when such figures are read as evidence that LLMs capture *people*. The remedy is to use Family 4b measures of computing the correlation within each question across matched people.

Second, Family 3 has a high, information-free floor on bounded scales. Although Family 3 requires matched individual records, nothing in the metric requires the prediction itself to be individuated. On a bounded scale, a constant answer is never far from any single response. If responses were spread uniformly across a rating scale, a benchmark that guesses the scale midpoint for everyone would already achieve an individual-level accuracy of about 0.75. For example, in the Twin-2K-500 mega-study (Toubia et al. 2025), twins prompted with each person's complete 500-question record achieve an accuracy of 0.748. "Twins" given no individual information at all score 0.734, and twins given only 14 demographic variables score 0.746, statistically indistinguishable from the full persona. In other words, essentially the entire headline "75% accurate" is available with very little data, or none.

The same floor problem takes a different form for Family 3 hit rates on multiple-choice items as well. For Family 3 the information-free benchmark is not random responding but the modal-answer predictor. For each question, always answer the option most respondents chose, with the mode estimated based on the training data. Because survey responses often concentrate on a few options, such a predictor can prove quite accurate with no individual respondent record at all (Miller 2026). To address this floor problem the researcher can compute *balanced accuracy* in two steps (Brodersen et al. 2010). First, for each response option separately, calculate the share of the respondents who actually chose that option whose twin also chose it, which is that option's *recall*. Second, average those recalls across the K options, giving every option equal weight no matter how many respondents chose it. The equal weighting removes the floor. A twin that answers the most popular option for everyone has a recall of one on that option and zero on every other, so its balanced accuracy is exactly 1/K, and 1/K is also what uniform random guessing scores. Guessing the modal answer therefore earns no credit under this balanced accuracy metric. For continuous and rating-scale responses, where balanced accuracy is not defined, the within-question correlation of Family 4b plays the analogous role, as Family 4b is already anchored at an information-free floor in a way that Family 1 distance measures are not.

Third, dividing twin accuracy by human test–retest reliability converts Twin-2K-500's 72% holdout accuracy into "88% relative accuracy" (Toubia et al. 2025), and yields the widely quoted figure for interview-grounded generative agents (Park et al. 2026), whose raw accuracy of 65.7% on held-out General Social Survey items becomes 82.6% once divided by participants' own two-week self-consistency of 79.5%. Test–retest is the correct ceiling, as no simulator should be expected to beat the person's own reliability, but a ratio to the ceiling conceals the distance from the floor. In that same study, agents given nothing but demographics reach an accuracy of 74% on the identical normalized scale. Reporting both corrections together avoids this: expressing performance as $(\text{Accuracy} - \text{Accuracy}_{\text{chance}})/(\text{Accuracy}_{\text{retest}} - \text{Accuracy}_{\text{chance}})$ anchors the results at both ends, between information-free responding and the respondent's own consistency. Where retest reliability is unavailable, as it often is, the floor correction alone is still worth applying.

Fourth, well-known scales, classic heuristics-and-biases items, and General Social Survey questions are likely to exist in the training corpus. Hence, holdout performance for such questions may overstate performance for new questions. Note that leakage of information is more concerning for aggregate or across-questions measures like Families 4a, 2, and 1 than for individual measures like Families 3 and 4b.

Fifth, twins are systematically under-dispersed, answering within a narrower range than the humans they represent (Peng et al. 2026). Accuracy also varies systematically across respondents, with higher accuracy for more educated, higher-income, and politically moderate people (Peng et al. 2026). Neither pattern is visible in mean accuracy measures. Hence, heterogeneity distributions should be reported alongside aggregate accuracy measures.

Sixth, to examine how much value individual respondents add, one could run the following *wrong-person twin* test. Rebuild each twin from a different respondent's record (or alternatively permute the twins' existing answers across respondents within each question). Whatever a twin learns by knowing the population, the wrong-person twin learns as well, so the gap between the wrong-person twin and the correct-person twin is the person-specific information (Miller 2026).

Taken together, these first five aspects point to the relationship between the question being asked and the data the twin was built from. Leakage matters because a question already in the training corpus is answerable from priors alone. The information-free floor is high because bounded scales let population priors do most of the work. Cross-question correlations inflate because item-level structure is available without knowing anything about any individual. In each case, what determines whether a twin's answer is informative is not a fixed accuracy measure of the twin but the distance between the question and the grounding data.

We therefore suggest that accuracy is best understood not as a property of a twin but as a property of *a twin–question pair*. A twin that reproduces a respondent's brand preferences in a category it was grounded in may carry little individual information about that same respondent's political attitudes. Thus, a single accuracy figure for a twin-building method means little without a statement of how far the evaluation questions sat from the grounding data, which is one reason we see high variance in previous accuracy measures. If accuracy varies systematically with distance, and if distance can be estimated before any ground truth is collected, then the question "can this twin be trusted with this question?" can be answered. Section 4 proposes an approach to address that question.

**3.3 Matching the measure to the decision: what industry actually needs**

For practitioners, the relevant question is not "*how accurate are digital twins*?" but "*which accuracy is required for this decision*?" Three tiers cover most commercial use.

*Tier 1: Population-level reads*

Concept screening, average willingness-to-pay, demand curves, and estimating average treatment effects from A/B-type tests require only Family 2 validity, question by question, against a matched human benchmark. This is BIN's evaluation, and it is the right evaluation for this tier, which makes BIN's results the appropriate cautionary tale even at the most forgiving level of accuracy. Aggregate users should additionally check dispersion and not only means. As noted in Section 3.2, twins are systematically under-dispersed, and a share simulator built on a compressed preference distribution will mis-forecast even when average WTP is correct.

*Tier 2: Heterogeneity*

Segmentation, targeting, positioning, product-line design, and differentiated pricing require knowing which customers want what, and will need at minimum Family 4b sorting plus Family 2 comparisons computed within segments. BIN's exploration of segmentation along low- and high-income groups finds evidence that is considerably worse than the aggregate accuracy measures. The differential accuracy across respondent groups described in Section 3.2 is particularly important here. For a targeting application, it is arguably worse than uniform inaccuracy, because the tool is most wrong exactly where segment comparisons are made.

*Tier 3: Individual-level uses*

Personalization, twin-as-panelist designs, respondent-level imputation, and within-person longitudinal simulation require Family 3 and/or Family 4b accuracy measures. Importantly, as mentioned earlier, Tier 3 accuracy satisfies, by definition, the use cases of Tiers 1 and 2 as well.

Overall, aggregate measures are often what industry needs to act on, and Tier 1 validation may legitimately suffice for a Tier 1 decision. But individual-level measures are what determine whether the instrument can be trusted beyond the exact questions on which it was benchmarked, whether it can be extended to segments, and how much human sampling it can actually replace. The three tiers describe what a decision requires, not what the researcher should report. Where matched individual-level data exist, all four families can be computed from the same data at essentially no extra cost, since they differ in how they aggregate rather than in what they need. The recommendation is therefore to compute every family the data support. Firms should demand to know which tier a vendor's accuracy claim was established at and should treat a Family 1 statistic as validation only for a Family 1 estimand. Used within its own tier, it is legitimate and inexpensive. However, when applied to support individual- or segment-level claims, it is misleading.

### 3.4 Recommendations

We close this section with a set of recommendations for evaluating digital twins, applicable to academic studies and commercial deployments alike.

- State the decision and its estimand first, then identify the accuracy family it requires. Report every family the data support, not only the one the use case demands. Do not apply a coarser family as evidence for a finer use. A cross-item correlation does not license claims about individuals or segments.
- Calculating within-question correlations across matched people and aggregating them (possibly via Fisher's z) is often a good measure for assessing whether twins preserve relative differences across people. Correlations across questions (Family 4a) are not the same as correlations across individuals (Family 4b). Make sure to understand the correlation reported.
- Report levels and sorting together. Individual-level accuracy (Family 3) and within-question correlation (Family 4b) answer different questions, and either alone can mislead. Their combination is preferable to Family 4a.
- Compare every accuracy number to an information-free floor and a test–retest ceiling. Report random-response, constant-midpoint, empty-persona, and demographics-only baselines (e.g., Toubia et al. 2025), and express performance as the increment above such baselines. If possible, calculate test–retest reliability as a ceiling to performance of the digital twins. For multiple-choice items, use the balanced accuracy measures (Brodersen et al. 2010; Miller 2026), which account for the modal-answer predictors.

- Use wrong-person twin and compare it to correct-person twin to assess the value of individual level information.
- Validate out of distribution**.** Use novel, pre-registered stimuli to limit training-data leakage. Research and firms that demonstrate accuracy on items that exist in the training data may overestimate the accuracy for truly novel findings (Blanchard et al. 2025). More broadly, as argued above, accuracy is a property of the twin–question pair rather than of the twin alone. Hence, any validation exercise is interpretable only alongside a statement of the distance between the focal question and training data.
- Report performance heterogeneity across respondent groups when evaluating segment-level use cases.
- For aggregate deployments, verify signs and dispersion, not just means.
- Do not attach conventional standard errors to Tier 1 and 2 synthetic estimates.

To summarize, the disagreement among published accuracy figures largely disappears when one considers the accuracy measure used. Papers reporting 90%+ accuracy are often measuring aggregate responses to items, papers reporting ~75% accuracy are often measuring a floor-inflated statistic, and papers reporting underwhelming performance are often measuring individuals against proper baselines on novel questions. The standard we propose asks future work, academic and commercial, to be explicit: establish the individual bar when you can, and when you cannot, state plainly which lower bar was met and for which decisions it suffices.

## 4. From *Whether* Digital Twins Work to *When* They Do

### 4.1 The forgotten question: Using LLMs to augment existing studies

Market research studies are designed under time and budget constraints. The common sequence is the following: designing the study to be executed, possibly combining qualitative and quantitative work, followed by fielding the study, analyzing the results, and reporting to the client. In this process, the study design must be frozen before fieldwork begins.

This process leads to a very common problem. It is not unusual for a client, or the research firm itself, to realize only after data collection has ended that an important question was never asked: an attribute left out of a conjoint design, a driver of choice that only became salient once early results were reviewed. We refer to this as the *forgotten question*. In conventional practice, answering it requires re-fielding, which is costly and slow, risks panel fatigue or attrition, and introduces comparability problems between waves. The forgotten question is thus a setting in which even an imperfect, low-cost estimate has substantial practical value.

Recent evidence suggests that this is also precisely the setting in which LLM-based simulation of survey respondents is most reliable. BIN evaluated LLM-based conjoint studies against parallel human studies and found that "off-the-shelf" LLM responses are too unreliable to act on. Baseline estimates of willingness-to-pay were at times overestimated by a factor of two to three, wrong-signed for brand and screen-size attributes and for new product flavors, and unstable across model versions. Similar results have been documented elsewhere as well (Bisbee et al. 2024; Motoki et al. 2024; Brucks and Toubia 2025; Toubia et al. 2025; Peng et al. 2026). However, BIN found that when the LLM is grounded in human data from the same category and population, it extrapolates well to a single new attribute that was withheld from the training data. Specifically, they found that fine-tuning on an existing toothpaste conjoint study flipped the incorrectly signed WTP estimates for two never-before-seen flavors to match the human benchmark, and fine-tuning on an existing laptop study moved the LLM's threefold overestimate of WTP for a built-in projector to match the human sign and approximate the magnitude. The same grounding did not transfer across categories (laptop data applied to tablets worsened alignment).

The forgotten question is very similar to these use cases explored by BIN. By construction, the researcher facing a forgotten question already possesses a complete human study on the relevant population and category. What is missing is one marginal question. Rather than asking the LLM to simulate an entire study, which, as BIN and others (e.g., Toubia et al. 2025; Peng et al. 2026) demonstrate is not sufficiently reliable for practical use, we ask it to leverage the full richness of a real respondent's data within the topic of the focal study and extend it by a single increment. In the context of digital twins, this means that, for each of the study respondents, an LLM is conditioned on that respondent's complete survey record and asked to answer the question the study never posed.

Establishing that single-question augmentation is a useful use case for digital twins requires answering at least two questions. First, do digital twins achieve satisfactory levels of accuracy when the task is not simulating an entire study but augmenting one related question? Second, which forgotten questions can be answered, and which cannot? Some forgotten questions sit close to the information contained in the original study. A question about price sensitivity, for example, is well supported by a study rich in cost and value attitudes. Others are simply too far from the data at hand, and for those, the twin is likely to fall back on the generic priors of its pretraining data, producing answers that may look plausible at the aggregate level but carry little respondent-specific information.

This second question is made evident by a finding of the BIN paper. When their baseline LLM estimates failed, a researcher without a human benchmark would have had no way to identify the failure in advance. The forgotten question is, by definition, a question for which no human benchmark exists. That is the entire reason to ask the LLM in the first place.

What is needed, therefore, is an *ex-ante* answerability diagnostic that measures whether a given forgotten question falls within the informational reach of the study at hand before moving to field a human study. The need for such a diagnostic is increasingly recognized. Ye and Yoga Narasimhan (2026) develop a question-level index, *rectification difficulty*, and predict it for new questions via meta-learning on a historical archive of paired human–LLM responses, in order to allocate a fixed human-response budget across the questions of a new survey. In what follows, we propose and empirically test a statistical approach for such a diagnostic.

### 4.2 Screening questions using the twins' own fit

When a digital twin answers a question, all of the respondent's information is included in its context, but the twin is a black box: we do not observe which variables it used, with what weight, or whether it used the provided data at all rather than its internal training knowledge.

We propose an approach to open this black box. For each candidate question, we feed into the prompt the respondent's responses to all of the survey questions apart from the forgotten question (which in practice we wouldn't have anyway) and record the twin's predicted answers across all respondents. We then fit a machine learning model, in our case, a random forest (Liaw and Wiener, 2002), in which the dependent variable is the twin's prediction for the forgotten question and the independent variables are the complete set of data points that were provided to the twin in the prompt (the responses to the questions in the survey that was fielded).[7] We use a random forest rather than a linear model because it accommodates complex non-linear relationships, aggregates many trees into a consensus prediction, and still yields interpretable statistics: an out-of-bag $R^2$ and variable importances. The approach can also be implemented with alternative models such as neural networks. We measure out-of-bag $R^2$ as $R^2 = 1 - \frac{MSE(Y)}{Var(Y)}$ similar to the way $R^2$ is calculated in OLS, but in the case of random forest measured on the out-of-bag sample to avoid overfitting (Fan et al. 2026). In the language of Section 3.2, the $R^2$ is an estimate of distance. It asks how much of the twin's answer is generated by the grounding data rather than supplied by the model's priors.

[7] The proposed approach is not limited to continuous scale closed-ended questions. When the forgotten question is binary or multiple-choice, the twin's prediction is a category, and the appropriate version of the diagnostic is a classification forest scored with a prevalence-robust measure such as out-of-bag balanced accuracy in place of $R^2$, for the reasons set out in Section 3.2. Open-ended responses that were collected in the fielded study can be embedded and entered on the right-hand side alongside the closed-ended items.

We then use the model's $R^2$ fit measure to determine whether the additional question can be answered by the twin. Crucially, the $R^2$ is computed from the twins' inputs and outputs alone. That is, human responses to the forgotten question are not used in this process, so the diagnostic can be computed for a forgotten question without collecting ground-truth data about that question, unlike hybrid estimation approaches that combine LLM predictions with human labels on the target question (Angelopoulos et al., 2023; Ye and Yoga Narasimhan, 2026).

The logic of the diagnostic is as follows. The model's fit measure ($R^2$) captures the share of variance in the twins' predictions that is systematically explained by the respondent's data. A high $R^2$ value means the twins' answers are meaningfully a function of the individual-level inputs. That is, the LLM is using the individual data it was given to construct the twin. This is a necessary condition for its answers to be respondent-specific rather than a repetition of population-level priors. A low $R^2$ means the twins' answers do not vary systematically with the respondent data. That is, the LLM is answering from its own general knowledge, and its output should not be trusted as a simulation of these particular respondents.[8]

Of course, our $R^2$ fit measure is a necessary but not sufficient condition, as it is possible that the LLM leverages its own data well to predict the outcome, but it predicts it incorrectly. Hence, next we explore this condition empirically for a specific empirical case study. As a by-product, the researcher can also obtain variable importances from the random forest algorithm to better understand which questions in the digital twin data help answer the additional question. For example, a forgotten question about price should load on cost- and value-related questions.

This diagnostic is related to the adaptive transfer rule of Fan et al. (2026), who use training error on the synthetic system to decide whether to apply a calibration mapping. Two differences are worth noting. First, our regressors are the respondent's actual survey record, the inputs supplied in the prompt, rather than the twins' responses to other questions, so the diagnostic speaks directly to whether the twin individuates on the data it was given. Second, our screen is a go/no-go decision about whether to pose the question to the twin at all, rather than a choice between two estimators.

[8] If the goal of the study is to obtain aggregate responses, or the focal question has low variance across respondents (e.g., everybody provides the same answer to the question), using the LLM's general knowledge may suffice.

### 4.3 Empirical application

To empirically test whether screening by the twins' own fit works, we use data from a large-scale survey collected by the Center for Customer-Based Execution and Strategy (C-CUBES) at the Jones Graduate School of Business, Rice University (Mittal and Tsiros 2025). The survey, fielded by TRC Insights, a Philadelphia-area marketing research company, in 2025, measured the importance customers place on value drivers across 18 for-profit and not-for-profit sectors spanning the U.S. consumer experience (from healthcare and education to banking, groceries, and law enforcement).

The sample comprised 3,063 respondents, drawn from a nationally representative panel. Each respondent rated up to three of the 18 sectors, yielding roughly 500 responses per sector. Additionally, respondents answered 20 demographic and background questions, covering characteristics such as education, marital status, and living situation, along with 108 attitude questions (value drivers) spanning the 18 categories, with 6 attitude questions per category, each rated on a 1–10 importance scale (1 = "Not at all important"; 10 = "Extremely important"). See Mittal and Tsiros (2025) for full details about the survey.

We validate the diagnostic with a leave-one-out design in which the "forgotten" question is one of the attitudinal questions and the twin is conditioned on (a) only the demographic questions, or (b) the demographic questions plus the remaining 5 attitudinal questions from the same category.[9] We asked the LLM (GPT-5.4-mini) to predict the answer for each respondent and for each of 108 attitudinal questions, one at a time, based on a prompt that included the respondent's answers to the other survey questions (either the demographics only, or the demographics plus the other attitude questions from the same category). See the exact prompt and model specification in the Web Appendix. We then estimated, for each held-out question, a random forest by fitting 500 decision trees where each tree is trained on a random bootstrap sample with approximately 63% of the data used for training and 37% used to calculate out-of-bag $R^2$ (see Web Appendix for details).

Using the respondents' actual responses to the forgotten questions, we can assess the accuracy of the digital twins and whether the $R^2$ measure is related to the ability of the digital twin to correctly predict the actual responses. Following the earlier discussion around accuracy measures, we use three accuracy measures. Each corresponds to one of the families defined in Section 3.1: individual-level correlation is a Family 4b measure, individual MAE is a Family 3 measure, and aggregate MAE is a Family 2 measure. We deliberately do not report Family 4a or Family 1 statistics. Both would look considerably more favorable than the measures below, and neither is the relevant bar for a use case whose value depends on respondent-specific information.

[9] We include only questions from the specific focal category of the forgotten questions since each respondent answered only three out of the 18 categories, making the cross-category data incomplete at the individual respondent level.

1) *Individual-level correlation* (Family 4b). For each held-out question, we compute the correlation between the twins' predicted answers and the respondents' actual answers across all individuals in the study. Unlike Family 4a correlations, this individual-level correlation captures whether the twins recover not just the average response, but *which* respondents rate the attribute higher or lower.
2) *Individual MAE* (Family 3). For each question, we average the absolute gap between each twin's answer and its own respondent's answer, measuring accuracy at the level of the simulated person. We do not normalize this measure since all questions are on the same scale.
3) *Aggregate MAE* (Family 2). This measure compares the mean of the twins' answers to the mean of the actual human answers, measuring the accuracy of the topline estimate a client would ultimately see.

Looking at Table 1, we can see that when using only the demographic questions, across all 108 attitude questions on a leave-one-out basis, the average correlation between the true individual-level responses to the forgotten question and the digital twin responses is a mere $r = 0.08$. Thus, in our setting, the comparison to the demographics-only condition mimics an information-free floor we discussed in Section 3.2. However, when we also include the topic-relevant questions, the correlation increases to $r = 0.57$. This level of accuracy is already meaningful, and it echoes the central finding of BIN: when digital twins were fine-tuned using relevant questions from the same category as the held-out question, the extrapolation worked.

The individual MAE shows the same pattern, improving from 2.01 to 1.46 scale points when the attitudinal questions are added. Interestingly, the aggregate MAE is better (lower MAE) for the lower information condition (Demographics only - 0.39) than the information-rich condition (demographics + attitudes 0.48).

**Table 1: Digital twin accuracy across the 108 attitude questions on a leave-one-out basis.**

| Digital twin | Individual correlation | Individual MAE | Aggregate MAE | Min. correl. | Max. correl. | % Ind. correl. > 0.5 |
|---|---|---|---|---|---|---|
| Demog. only | 0.08 | 2.01 | 0.39 | −0.15 | 0.29 | 0.0% |
| Demog. + attitudes | 0.57 | 1.46 | 0.48 | 0.27 | 0.80 | 74.1% |

*Notes: Each row summarizes the 108 held-out attitude questions within the stated information condition. Correlation is between twin-predicted and actual respondent answers at the individual level (Family 4b); MAE is in scale points.*

While the overall correlation is reasonable, for practical purposes, a correlation of 0.57 between the true responses and the digital twin responses may be below the bar for what a market research team or a client is willing to accept. More importantly, even when using the attitudinal questions to generate the digital twins, 25.9% of the questions had a correlation coefficient smaller than 0.5, and the question with the minimum correlation had a modest correlation coefficient of 0.27, which would be below the bar for almost any meaningful exploration. This raises the question of whether the random forest $R^2$ can help select the questions with higher correlations between the true responses and the digital twin responses.

First, looking at the $R^2$ of the two information conditions, we find that, indeed, demographic information does not carry much information in explaining the digital twin responses (average $R^2$ across the 108 questions = 0.14). Consistent with the findings of BIN, this means that when we provide only demographic information, the LLM relies mostly on external information, if at all, to provide the individual answers. On the other hand, when we use the demographics plus the attitudinal questions, the average $R^2 = 0.65$, demonstrating heavy reliance on the information provided in the digital twin prompt.

Second, within the demographics-plus-attitudes condition, we find correlations of $r = 0.61$ ($p < .001$), $r = -0.22$ ($p = 0.024$), and $r = -0.37$ ($p < .001$) between the random forest $R^2$ and the individual-level correlations, the aggregate MAE, and the individual MAE, respectively (note that negative correlations are expected with the MAE measures, since a lower MAE means higher accuracy). See Figure 1 for the scatter plot of the relationship between the random forest $R^2$ and the individual correlation accuracy metric across the 108 questions. The pattern suggests that the $R^2$ fit measure carries information about which questions can be answered with a higher level of accuracy.[10]

[10] The two outlier questions with very high $R^2$ but low correlations are questions about safety and diversity, equity, and inclusion, which may reflect the stereotyping and ideology bias reported in Peng et al. (2026).

**Figure 1. Scatter plot of the random forest $R^2$ screen versus the individual-level correlation accuracy metrics (demographics + attitudes condition, 108 questions).**

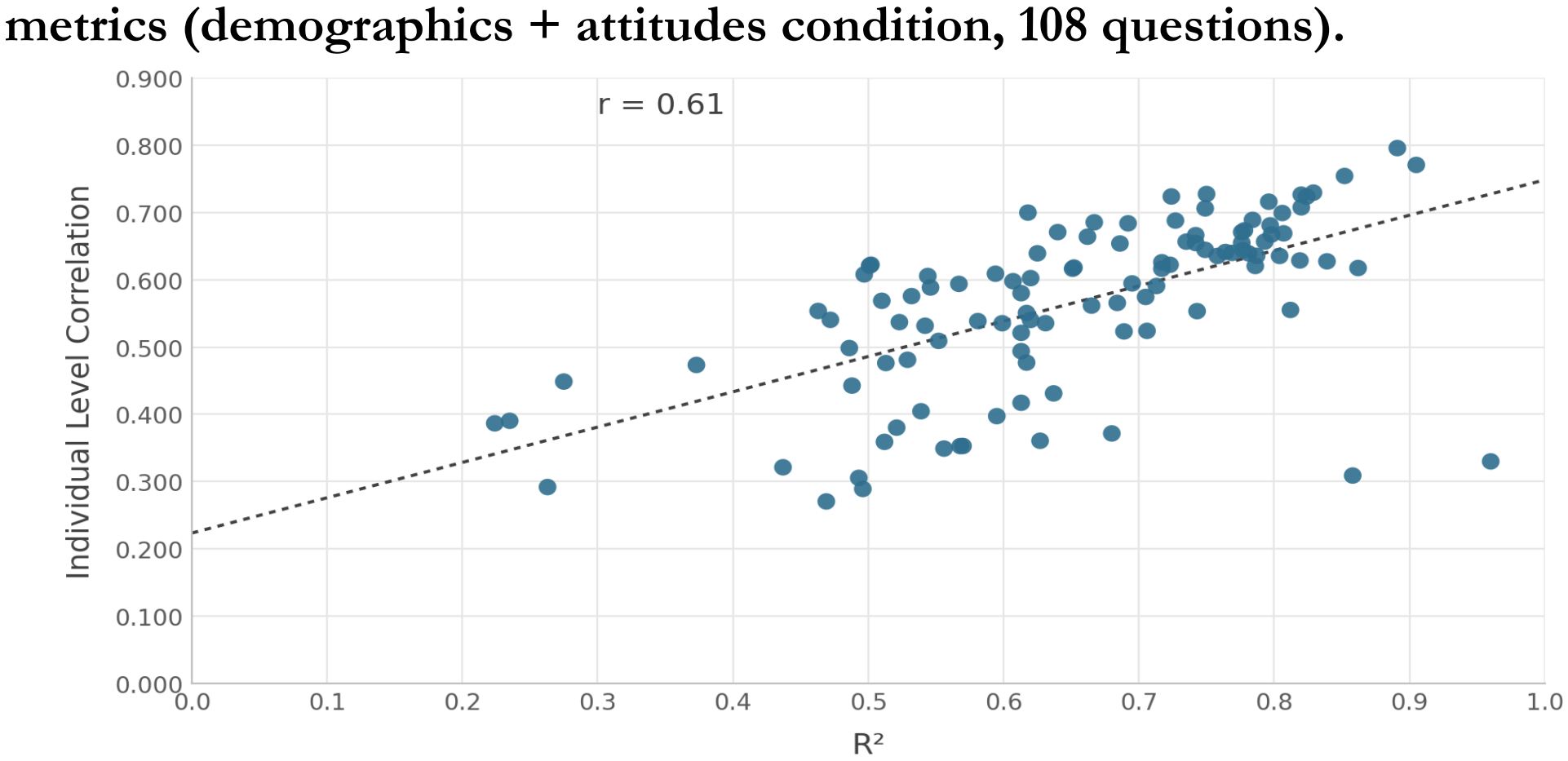

### 4.3.1 Using the $R^2$ metric as a screen

Next, we explore whether the $R^2$ metric can be used as a screening mechanism for deciding whether a forgotten question should be answered by the digital twin in the first place.

Table 2 reports what happens when we restrict attention to questions whose random forest $R^2$ values clear higher thresholds successively, within the demographics-plus-attitudes condition. With no screen, as we reported in Table 1, the average individual-level correlation between twin and human answers is 0.57, and 25.9% of questions fall below a correlation of 0.50, a failure rate that would be unacceptable in practice, particularly because the failures are not identifiable from the twins' output alone. Requiring $R^2 > 0.7$[11] retains 43% of the candidate questions while raising the average correlation to 0.65 and, more importantly, cutting the share of poorly answered questions to only 4.3%. The individual-level MAE shows a similar pattern.

Looking into the specific questions that the $R^2 > 0.7$ screen rejects, we indeed see that the least answerable items are systematically those least anchored in the attitudinal data provided (in our data, most notably the diversity, equity, and inclusion importance items, whose twin answers had low actual accuracy across categories and unreliable $R^2$ accuracy measures). The random forest attribute importances provide a window into the information used to predict the forgotten question. When the forgotten question is, say, the importance of streaming-service affordability, the twin's answer is driven overwhelmingly by the respondent's stated importance of adjacent attributes in the same category (content selection, streaming quality), with demographics contributing little. This is exactly the pattern one would expect if the twin were reasoning from the respondent's record.

[11] In practice, the $R^2$ cut-off level should be estimated using a hold-out sample.

**Table 2. The $R^2$ screen: accuracy of twin answers to held-out questions, by random forest $R^2$ threshold (demographics + attitudes condition, 108 questions).**

| Screen | Questions retained | Mean twin–human correlation | Share of questions with r < 0.50 | Mean ind. MAE | Mean Agg. MAE | Attainable $R^2$ |
|---|---|---|---|---|---|---|
| No screen | 108 (100%) | 0.57 | 25.9% | 1.46 | 0.48 | 0.542 |
| $R^2 > 0.5$ | 94 (87%) | 0.59 | 18.1% | 1.43 | 0.48 | 0.549 |
| $R^2 > 0.6$ | 72 (67%) | 0.62 | 11.1% | 1.33 | 0.42 | 0.561 |
| $R^2 > 0.7$ | 46 (43%) | 0.65 | 4.3% | 1.19 | 0.32 | 0.575 |

*Notes: Each row restricts the 108 held-out questions to those whose random forest $R^2$ — the twin's prediction regressed on all variables provided to the twin — exceeds the stated threshold. Correlation is between twin-predicted and actual respondent answers at the individual level (Family 4b); MAE is in scale points.*

To gauge how much of each answer was recoverable from the record at all, we repeat the random forest and out-of-bag $R^2$ from regressing the *true* response to the forgotten question (one at a time) on the demographics and category-specific responses (Toubia et al. 2025, Web Appendix A2.3). This attainable $R^2$ (reported in the last column of Table 2) requires ground truth and is therefore not usable as a screen, but its square root reflects the attainable level of information that one can extract from the training data provided for the digital twin. It averages 0.542 across the 108 questions, implying a ceiling correlation of 0.736, so the observed digital twin correlation with the true responses of 0.57 is 0.57/0.736=77% of what was attainable, rather than 57% relative to a perfect correlation of 1. The comparison also shows the value of the screen. As we increase the cutoff for the screening we are doing two things: 1) We select questions that may be easier to answer given the data, and 2) we are selecting questions for which the LLM digital twin is better able to extract information from the training data. The rise of the attainable $R^2$ from 0.542 to 0.575, captures the former, while the fact that the extraction efficiency rises from 77% to 86% (0.65/sqrt(0.575)=0.86) captures the latter.

### 4.3.2 Replacing the $R^2$ screening with Question-Embedding Similarity

One limitation of the $R^2$ statistical screening approach is that it requires running the digital twins for each forgotten question to assess its likely accuracy level. This can be costly in terms of both running time and compute cost. A faster and cheaper approach to measure the distance between the question being asked (the forgotten question) and twins' training data can be to calculate how semantically similar the focal question is to the data that comprise the twins. Theoretically, if the question is similar to the questions within the twin's input data, then we may expect a higher accuracy since there is relevant information in the twin's training data, and the degree of extrapolation needed to go from the twin training data to the new question is low.

To assess semantic similarity, we extracted the embedding representation for each question using text-embedding-ada-002. We then calculated the cosine similarity between the embedding representation of each question and the forgotten question. For each attitudinal question, we calculated the max cosine similarity between that specific question and every other question in the twin's training data.[12] The max cosine similarity metric had a correlation of $r = 0.53$ ($p < 0.001$) with twin-human correlation as well as a correlation of $r = 0.31$ ($p < 0.001$) with random forest $R^2$. Similar to the random forest $R^2$ results in Table 2, the accuracy metrics improve as max cosine similarity increases (see Table 3), but the increase is limited if we only consider questions with a maximal similarity above 0.75. While computationally cheap and less time-intensive, the gains are smaller than the random forest $R^2$ (with the exception of the individual MAE measures) and are not immediately realized with each cutoff.

An important difference between the two screens is that the $R^2$ screen asks whether the twins' input data carry enough information to differentiate respondents, whereas the semantic similarity screen asks only how closely the forgotten question resembles the questions already fielded. The latter is computed from question wording alone and is identical for every respondent, so it speaks to question-level answerability rather than respondent-level differentiation. In that sense the two screening rules are complementary.

**Table 3. The semantic similarity screen: accuracy of twin answers to held-out questions, by maximal cosine similarity threshold (demographics + attitudes condition, 108 questions).**

| Screen | Questions retained | Mean twin–human correlation | Share of questions with r < 0.50 | Mean individual MAE | Mean Aggregate MAE |
|---|---|---|---|---|---|
| No screen | 108 (100%) | 0.57 | 25.9% | 1.46 | 0.48 |
| Max > 0.65 | 102 (94.4%) | 0.58 | 22.5% | 1.40 | 0.49 |
| Max > 0.70 | 78 (72.2%) | 0.59 | 17.9% | 1.27 | 0.43 |
| Max > 0.75 | 42 (38.9%) | 0.63 | 14.3% | 1.17 | 0.40 |

*Notes: Each row restricts the 108 held-out questions to those whose embedding maximal cosine similarity between the focal questions and questions with the highest similarity in the twin training data exceeds the stated threshold. Correlation is between twin-predicted and actual respondent answers at the individual level (Family 4b); MAE is in scale points.*

---

[12] Theoretically using the max cosine similarity between the focal held-out question and every other question in the twin's training data is sufficient, because even a single very relevant question in the training data should guide the twin how to respond to the focal held-out question. Empirically, we also test a version with the average cosine similarity between the focal held-out question and the other questions in the twin's training data and found the maximum similarity to perform better.

### 4.3.3 Replacing the $R^2$ screening with practitioner judgment

Going back to the opening example of this article, the question we ask is whether a skilled researcher could serve as a screener as well. In other words, can a skilled researcher predict whether the digital twin is capable of answering a forgotten question well?

To answer this question, we surveyed N = 16 marketing researchers at TRC Insights, spanning research, account management, and analytics roles. We focused on a subset of 18 of the 108 attitudinal questions, covering three categories (health insurance, non-cash payment services, and streaming services), and evaluated each under the demographics plus remaining attitudes condition. The researchers were shown the forgotten question and the information the digital twin would receive (the demographics plus the remaining 5 attitude questions in that category) and asked to rate, on a 0–100 scale, their confidence that the digital twin could recover respondents' answers. For each question, we averaged the responses across the 16 experts. The average inter-rater correlation across questions was 0.816.

The experts' predictions were informative. Average expert confidence correlated with the twins' realized accuracy at $r = 0.77$ ($p < 0.001$) across the 18 questions and with the random forest $R^2$ at $r = 0.87$ ($p < 0.001$). That is, the human screen and the statistical screen largely agree on which questions are answerable. As Table 4 shows, expert confidence in the digital twin's predicted accuracy functions as a usable screening mechanism as well. Confidence thresholds between 55 and 65 retain 61%–83% of candidate questions while raising the mean twin–human correlation from 0.60 to 0.65–0.67 and cutting the share of poorly answered questions from 27.8% to 7%–13%. With only 18 questions, these correlations are imprecisely estimated. Hence, they should be taken with caution.

**Table 4. The expert screen: accuracy of twin answers to held-out questions, by expert confidence threshold (18 questions).**

| Screen | Cases retained | Mean twin–human correlation | Share of cases with r < 0.50 | Mean individual MAE | Mean Aggregate MAE |
|---|---|---|---|---|---|
| No screen | 18 (100%) | 0.60 | 27.8% | 1.36 | 0.37 |
| Confidence > 55 | 15 (83.3%) | 0.65 | 13.3% | 1.17 | 0.35 |
| Confidence > 60 | 14 (77.8%) | 0.67 | 7.1% | 1.11 | 0.32 |
| Confidence > 65 | 11 (61.1%) | 0.66 | 9.1% | 1.12 | 0.31 |

*Notes: Confidence is the mean rating (0–100), across surveyed experts, that the twin could recover respondents' answers to the held-out question given the stated information set. The 18 cases comprise six attribute-importance questions in each of three categories (health insurance, non-cash payment services, and streaming services), under the demographics-plus-attitudes condition.*

Taken together, these results suggest a meaningful opportunity for the forgotten question problem. When a client identifies a question the study never asked, the research firm can assess its answerability before generating, let alone acting on, any twin-based estimates, using either the judgment of its own experienced researchers, the $R^2$ diagnostic, or both in sequence. Questions that clear the screening can be answered by the twins at a small fraction of the cost and turnaround time required for re-fielding, with an empirically grounded expectation of accuracy. Questions that fail the screening are flagged as requiring new human data. This preserves the division of labor recommended by BIN — human data first, LLM extrapolation second — while providing a way to know in advance not only *whether* we should use digital twins but, more importantly, *when* the extrapolation can be trusted.

## 5. General Discussion

The debate over synthetic data has largely been conducted as a referendum on a yes-or-no question: can LLMs replace human respondents? We have argued throughout that this is the wrong question (Sambandam and Netzer 2025), and that two better questions would be: 1) which standard of accuracy does a given decision require? and 2) which questions fall within the informational reach of the data already in hand? In this paper, we provide a conceptual framework to answer these questions and initial empirical evidence.

BIN provide a pair of findings that point in opposite directions and are more useful together than either is alone. Off-the-shelf LLM responses are unreliable in ways a researcher cannot detect without a human benchmark, and grounding the model in human data from the same category and population can help to improve the accuracy of synthetic data for a single adjacent attribute but does not transfer across categories and does not recover heterogeneity. While these results may seem discouraging, they may provide a description of the conditions under which synthetic data can be trusted: same population, same category, small extrapolation. Stated in the terms we have used throughout, those three conditions point to small distance between the question being asked of the twin and the grounding data.

### 5.1 Beyond the Forgotten Question: The Forgotten Step

Our empirical work addressed the forgotten question: a single item omitted from a study that has already been fielded. The same logic extends to a larger and, we suspect, more valuable target, which is the forgotten step.

To examine a question rigorously, academic research and large public data collection efforts such as those of the Census Bureau proceed through many sequential stages, often at considerable expense and over long horizons. In commercial marketing research, this almost never happens. Given commercial budget and time constraints, a study is typically one step, occasionally two. Qualitative exploration is compressed or skipped. Price ranges are set by judgment rather than tested. Claim wordings go into the field without a pretest. Attribute lists are finalized in a meeting rather than derived from data. Critically, the steps that get cut are not the final study, which is the deliverable the client is paying for. They are the preparatory steps that would have made that final study better.

This is where synthetic data can help (Arora et al. 2025; Korst et al. 2026), and it is worth noting that BIN's own results are more encouraging for this use than for the one they set out to evaluate. Their baseline conjoint estimates failed as substitutes for a study, but they were frequently in the right ballpark on sign and ordering, which is the standard a pretest has to meet. The forgotten steps are precisely the low-stakes, high-iteration, directionally oriented tasks for which segment-level personas or even ungrounded LLM responses may be adequate. A researcher can screen twenty claim wordings down to five, bound a plausible price range, check whether an attribute list is reasonable, or pressure-test a questionnaire for order and framing effects (Brucks and Toubia 2025), at a cost and speed that makes the exercise worth doing. The human study then goes into the field better specified than it otherwise would have been.

A pretest that is directionally right is useful even when it is imprecise, because the output of such an LLM pretest is to improve the research process or help to create a better research instrument rather than a business decision. Errors are caught by the human study that follows rather than propagating into the decision-making. Framed this way, synthetic data are not competing with human data for the same budget. They can restore steps that the budget had already eliminated. This preserves scarce and expensive human samples for the consequential final study while expanding the total amount of learning that surrounds it (Sambandam and Netzer 2025). It also sidesteps the polarization with which we opened. Neither the claim that human respondents are obsolete nor the claim that synthetic respondents have no legitimate place speaks to a use case in which nothing is being replaced.

### 5.2 Heterogeneity and Segment-Level Uses

At the heart of many marketing research studies is correctly identifying heterogeneity across consumers. This is where we are less optimistic about much of the current use of synthetic data. The demographic and heterogeneity failure documented in the BIN paper is arguably the most important negative result in the paper for practitioners. The promise that most people attach to synthetic respondents is the ability to simulate segments cheaply. BIN show that even with fine-tuning, the model approximates the population means while failing to recover differences across income, gender, and political groups, and sometimes inverting them.

Our own results echo this from a different direction. In the demographics-only condition, the average individual-level correlation between twin and human answers was 0.08, and not one of the 108 questions cleared a correlation of 0.5. Yet the aggregate MAE in that condition was 0.39, slightly better than in the demographics-plus-attitudes condition. A twin that knows nothing about a person can still produce a topline a client would accept. Peng et al. (2026) add two further distortions: twins are systematically under-dispersed, and their accuracy varies across respondent groups, performing better for more educated, higher-income, and politically moderate respondents. For a targeting or positioning application, differential accuracy is worse than uniform inaccuracy, because the instrument is least reliable exactly where the segment comparison is being made.

This should serve as a word of warning against relying on Type 2 data for segment-level personas (provide segment characteristics to generate simulated segment behavior). The gap between what vendors often claim and what has been demonstrated is widest in such situations. Recovering heterogeneity requires individual-level grounding data along the lines of generating individual-level digital twins. While costly, there is no shortcut from a population prior to individual or segment differences.

### 5.3 The Role of Human Respondents

We close on a more speculative note. It is possible that we are being too conservative in imagining what digital twins change. Thinking expansively, one could imagine knocking out all the conventional pillars of the research workflow, including structured quantitative questionnaires, discussion guides, fixed data collection protocols, the separation between qualitative and quantitative instruments, and the segregated analytic toolkit, and designing the process fresh. Many of these conventions are artifacts of constraints that are rapidly ceasing to bind. Closed-ended scales exist partly because coding open-ended responses was expensive. Instruments are frozen before fieldwork partly because re-fielding was slow and costly, which is the very constraint that creates the forgotten question. Segmentation is a separate study partly because one could not ask everything of everyone. A research process designed today, with the ability to interrogate a respondent's full record after the fact, might not like the one we inherited.

One element, however, should not be reconsidered, and that is the involvement of the human respondent. In the marketing ecosystem, the individual consumer is the fulcrum around which the entire system revolves. A digital twin, however well-constructed, is a compression of what some person has already revealed. While it can rearrange evidence of preference, it cannot originate preference. Every twin is downstream of a human who answered something. The problem is not merely that a model's training corpus has a cutoff date, an issue that retrieval and continual updating may partly address. It is that a simulator has no independent channel through which new preference can reach it. When tastes shift, when a category is genuinely new, when a respondent does something that the model's prior knowledge did not anticipate, the information has to enter the system from a person. And it is worth recalling that these surprises are not incidental to the enterprise. As we noted at the outset, the client is not paying for the eighty percent that a seasoned researcher or a capable model could have predicted. The client is paying for the ten to twenty percent that no one knew before the study began. Regardless of progress in synthetic data development, it would be foolhardy to remove the thinking, feeling, adapting, and sometimes irrational consumer decision-maker from the system that exists to understand them.

The productive framing is therefore not human or machine but the combination of the two, in which each is applied where it is cheap and reliable and checked where it is not (Karlinsky-Shichor and Netzer 2024; Arora et al. 2025). Our expert-screening result is a small instance of the same principle. The statistical diagnostic and the experienced researcher's judgment largely agreed on which questions were answerable, correlating at 0.87, which means each can substitute for the other when the other is expensive, and each can validate the other when the stakes warrant both.

We began with a researcher in 2016 who declined to state the answer he already knew, because no one would act on it without data behind it. His 2026 counterpart can produce that answer in minutes. The interesting question is what he does with the time he has saved. Our answer is that he should spend it on the steps he never had the budget to run and on the respondents he could never afford to reach, and then still go to the field.

# Web Appendix

### A1. LLM Specification

We asked the LLM (GPT-5.4-mini) to predict the answer for each respondent and for each of 108 attitudinal questions, one at a time, based on a prompt that included the respondent's answers to the other survey questions (either the demographics only, or the demographics plus the other attitude questions from the same category). See the exact prompt in Appendix A2.

We used GPT-5.4-mini, accessed between 5/21/2026-5/31/2026. We set the effort level to low/minimal (limited testing on higher level of effort showed little improvement). Each question was asked as a single cell, one question and one respondent at a time, with no memory. Each response is based on a single call to the agent. All responses were within the requested range of 1-10 so no errors or missing data.

### A2. Sample prompt for the digital twin predictions

Below is a transcript where someone was asked questions and gave their answers. Pretending you are this person, please answer the following question: [On a scale of 1 to 10 where 1 means not at all important and 10 means extremely important, how important is the following to you when choosing or continuing to use a streaming service? Cost/affordability (monthly subscription price, no additional/hidden fees)]

*Followed by questions and answers to the training questions:*

Question 1:

Answer 1:

Question 2:

Answer 2:

….
Question N:

Answer N:

### A3. Random Forest Specification

In estimating the random forest to calculate the out-of-bag $R^2$ we left the parameters at their default setting which was:

- ntree = 500 (number of trees)
- mtry = p/3 (number of variables sample at each split where p is the number of variables in the regression)
- replace = true (sampling with replacement)
- sampsize = 0.632 * # of rows (the amount of data used for training with the remaining for oob validation)

- nodesize = 5 (maximum size of terminal nodes)
- seed - random seed set at default

For the Cross-validation or out-of-bag procedure, each tree was sampled with replacement as mentioned above (~63% training, 37% validation). $R^2$ is based on cumulative trees, so $R^2$ (1) is based on only the first tree, $R^2$ (100) is based on the first 100 trees and the final one, $R^2$ (500) is based on all 500 trees. For each observation, their predicted value is only based on the trees they were held out of

To test for effect of the random seed, we did test it a few different times to make sure we get consistent $R^2$.